\documentclass{article}
\usepackage{authblk}
\usepackage[utf8]{inputenc}
\usepackage{main}
\usepackage{microtype}
\usepackage{subcaption}
\usepackage{graphicx}
\usepackage{times}
\usepackage{latexsym}
\usepackage{amsmath}
\usepackage{float}
\usepackage{footnote}
\usepackage{enumitem}
\usepackage{bm}
\usepackage{arydshln}
\usepackage{booktabs}
\usepackage{array}     
\usepackage{multicol}
\usepackage{multirow}
\usepackage{color}
\usepackage{xcolor}     
\usepackage{colortbl}
\usepackage{bbding}
\usepackage{makecell}
\usepackage{mathtools}
\usepackage{imakeidx}
\usepackage{longtable}
\usepackage{wrapfig}
\usepackage{algorithmic}
\usepackage{rotating}
\makeindex
\usepackage{arydshln}
\usepackage{lipsum}
\usepackage{natbib}
\usepackage[toc]{multitoc}
\usepackage[edges]{forest}
\usepackage[normalem]{ulem}
\definecolor{mydarkblue}{rgb}{0,0.08,0.45}
\usepackage[colorlinks=true,linkcolor=mydarkblue,citecolor=mydarkblue,filecolor=mydarkblue,urlcolor=mydarkblue]{hyperref}
\usepackage{CJKutf8}
\usepackage{awesomebox} 
\usepackage{bbding}
\usepackage[most]{tcolorbox}
\usepackage{booktabs}
\usepackage{geometry}
\definecolor{wkblue}{rgb}{0.2, 0.3, 0.6}
\definecolor{meta-color}{rgb}{0.5, 0.5, 0.5}
\usepackage{amsmath}
\usepackage{enumitem}
\usepackage{lscape} 
\usepackage{booktabs}
\usepackage{algorithm}   
\usepackage{setspace}

\usepackage{algorithmic}
\usepackage{tabularx,booktabs}
\usepackage{makecell}

\usepackage{amssymb}
\usepackage{amsfonts}

\usepackage{booktabs} 
\usepackage{geometry} 

\usepackage{multirow}

\usepackage[tikz]{bclogo}
\usepackage[framemethod=tikz]{mdframed}
\definecolor{bgblue}{RGB}{245,243,253}
\definecolor{ttblue}{RGB}{91,194,224}

\usepackage{pgfplots}
\usepackage{pgfplotstable}

\usepackage{wrapfig}
\usepackage{graphicx}

\mdfdefinestyle{mystyle}{%
  rightline=true,
  innerleftmargin=10,
  innerrightmargin=10,
  outerlinewidth=3pt,
  topline=false,
  rightline=true,
  bottomline=false,
  skipabove=\topsep,
  skipbelow=\topsep
}

\newtcolorbox{myboxi}[1][]{
  breakable,
  title=#1,
  colback=red!5,
  colbacktitle=red!5,
  coltitle=black,
  fonttitle=\bfseries,
  bottomrule=0pt,
  toprule=0pt,
  leftrule=2pt,
  rightrule=2pt,
  titlerule=0pt,
  arc=0pt,
  outer arc=0pt,
  colframe=red,
}

\newtcolorbox{myboxnote}[1][]{
  breakable,
  title=#1,
  colback=orange!0,
  colbacktitle=orange!0,
  coltitle=black,
  fonttitle=\bfseries,
  bottomrule=0pt,
  toprule=0pt,
  leftrule=2pt,
  rightrule=2pt,
  titlerule=0pt,
  arc=0pt,
  outer arc=0pt,
  colframe=orange,
}

\newtcolorbox{myboxii}[1][]{
  breakable,
  freelance,
  title=#1,
  colback=white,
  colbacktitle=white,
  coltitle=black,
  fonttitle=\bfseries,
  bottomrule=0pt,
  boxrule=0pt,
  colframe=white,
  overlay unbroken and first={
  \draw[red!75!black,line width=3pt]
    ([xshift=5pt]frame.north west) -- 
    (frame.north west) -- 
    (frame.south west);
  \draw[red!75!black,line width=3pt]
    ([xshift=-5pt]frame.north east) -- 
    (frame.north east) -- 
    (frame.south east);
  },
  overlay unbroken app={
  \draw[red!75!black,line width=3pt,line cap=rect]
    (frame.south west) -- 
    ([xshift=5pt]frame.south west);
  \draw[red!75!black,line width=3pt,line cap=rect]
    (frame.south east) -- 
    ([xshift=-5pt]frame.south east);
  },
  overlay middle and last={
  \draw[red!75!black,line width=3pt]
    (frame.north west) -- 
    (frame.south west);
  \draw[red!75!black,line width=3pt]
    (frame.north east) -- 
    (frame.south east);
  },
  overlay last app={
  \draw[red!75!black,line width=3pt,line cap=rect]
    (frame.south west) --
    ([xshift=5pt]frame.south west);
  \draw[red!75!black,line width=3pt,line cap=rect]
    (frame.south east) --
    ([xshift=-5pt]frame.south east);
  },
}

\usepackage{fontawesome5}
\usepackage{fancyhdr} 
\usepackage{blindtext} 
\usepackage{makecell}

\DeclareCaptionFont{black}{\color{black}}

\definecolor{myblue}{rgb}{0.9, 0.1, 0.94}
\definecolor{mygreen}{rgb}{0.64, 0.56, 0.88}
\definecolor{myyellow}{rgb}{0.68, 0.6, 0.1}
\definecolor{fancygreen}{rgb}{0.33, 0.68, 0.20}
\definecolor{salmon}{rgb}{0.94, 0.52, 0.49}
\definecolor{tablegreen}{rgb}{0.82, 0.94, 0.75}
\definecolor{tableblue}{rgb}{0.81, 0.90, 0.94}
\definecolor{tablered}{rgb}{0.97, 0.85, 0.85}
\definecolor{tableorange}{rgb}{0.96, 0.85, 0.81}

\newenvironment{itemize*}%
 {\leftmargini=10pt\begin{itemize}%
  \setlength{\itemsep}{0pt}%
  \setlength{\parskip}{0pt}%
  }%
 {\end{itemize}}
\newenvironment{enumerate*}%
 {\begin{enumerate}%
  \setlength{\itemsep}{0pt}%
  \setlength{\parskip}{0pt}}%
 {\end{enumerate}}

\usepackage{xcolor}
\usepackage{listings}  

\newcommand\JSONnumbervaluestyle{\color{blue}}
\newcommand\JSONstringvaluestyle{\color{red}}

\newif\ifcolonfoundonthisline

\makeatletter

\lstdefinestyle{json}
{
  showstringspaces    = false,
  keywords            = {false,true},
  alsoletter          = 0123456789.,
  morestring          = [s]{"}{"},
  stringstyle         = \ifcolonfoundonthisline\JSONstringvaluestyle\fi,
  MoreSelectCharTable =%
    \lst@DefSaveDef{`:}\colon@json{\processColon@json},
  basicstyle          = \ttfamily,
  keywordstyle        = \ttfamily\bfseries,
}

\newcommand\processColon@json{%
  \colon@json%
  \ifnum\lst@mode=\lst@Pmode%
    \global\colonfoundonthislinetrue%
  \fi
}

\lst@AddToHook{Output}{%
  \ifcolonfoundonthisline%
    \ifnum\lst@mode=\lst@Pmode%
      \def\lst@thestyle{\JSONnumbervaluestyle}%
    \fi
  \fi
  \lsthk@DetectKeywords%
}

\lst@AddToHook{EOL}%
  {\global\colonfoundonthislinefalse}

\makeatother

\usepackage{etoolbox}
\usepackage{natbib}
\usepackage{url}
\newcounter{bibcount}
\makeatletter
\patchcmd{\@lbibitem}{\item[}{\item[\hfil\stepcounter{bibcount}{[\thebibcount]}}{}{}
\renewcommand\NAT@bibsetup%
  [1]{\setlength{\leftmargin}{\bibhang}\setlength{\itemindent}{-\parindent}%
      \setlength{\itemsep}{\bibsep}\setlength{\parsep}{\z@}}
\makeatother

\definecolor{mybrown}{RGB}{128,64,0}

\definecolor{titlecolor}{HTML}{4c9cff}
\definecolor{coolblue3}{rgb}{0.91, 0.94, 0.98}

\begin{document}



\title{MarathonPR: Bridging the Gap to Long-Horizon Agency\\ with Multi-PR Chains}

\title{daVinci-Agency: Unlocking Long-Horizon \\Agency Data-Efficiently}

\author[1,2,4]{Mohan Jiang}
\author[1,4]{Dayuan Fu}
\author[1,2,4]{Junhao Shi}
\author[2,4]{Ji Zeng}
\author[2,4]{Weiye Si}
\author[1,2,4]{Keyu Li}
\author[1,2,4]{Xuefeng Li}  
\author[3,4]{Yang Xiao}
\author[3]{Wenjie Li}
\author[1,2]{Dequan Wang}
\author[1,2,4]{Pengfei Liu\textsuperscript{†}}
\affil{SII \quad \textsuperscript{2}SJTU \quad \textsuperscript{3}PolyU \quad \textsuperscript{4}GAIR}

\newcommand{\MethodFull}{Chain of Pull Requests\xspace}
\newcommand{\methodShort}{daVinci-Agency\xspace}
\newcommand{\ourDataset}{daVinci-Agency\xspace}
  
\maketitle
\pagestyle{fancy}
\thispagestyle{fancy}
\fancyhead{}
\lhead{
  \raisebox{-0.3cm}{\includegraphics[height=0.95cm]{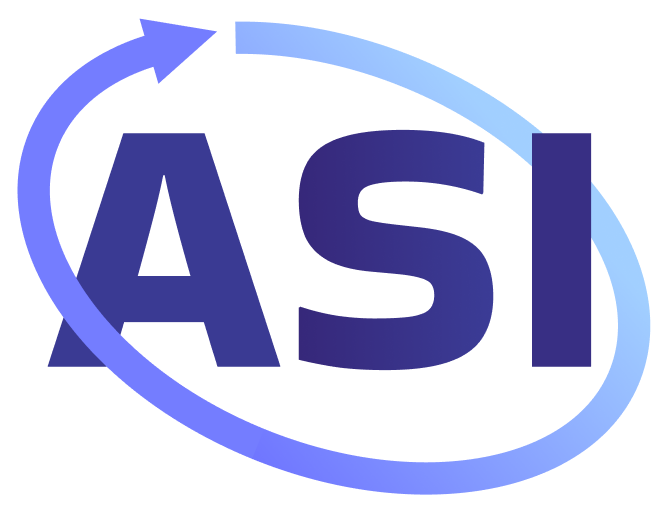}}
}
\rhead{%
  \raisebox{-0.2cm}{\includegraphics[height=0.7cm]{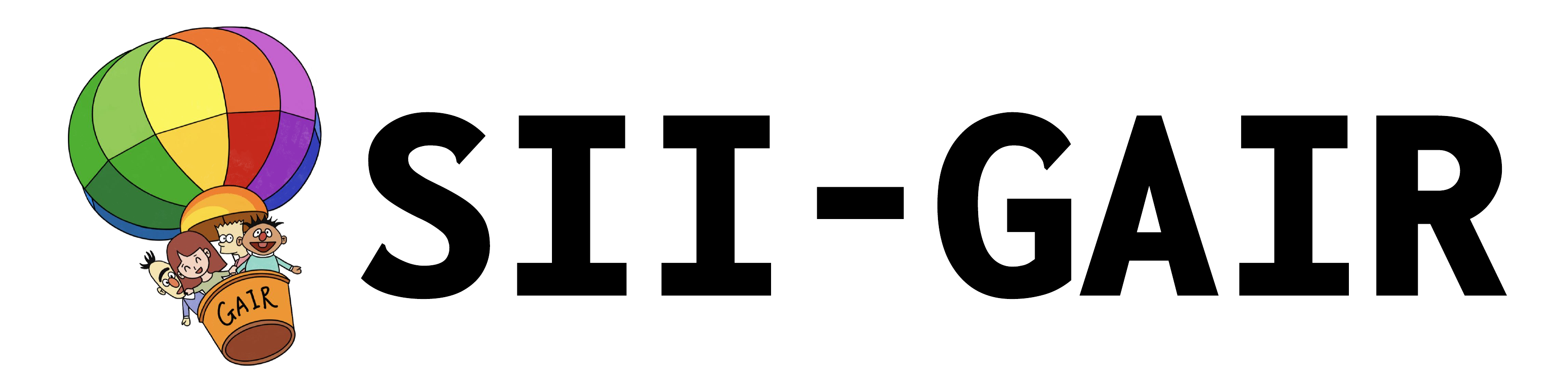}}%
}
\renewcommand{\headrulewidth}{0pt}
\setlength{\headsep}{2mm} 


\renewcommand{\thefootnote}{}
\footnotetext{† Corresponding author.}
\vspace{-30pt}
\begin{center}
    \quad \textbf{Open Source:} 
\quad \href{https://github.com/GAIR-NLP/daVinci-Agency}{\textcolor{black}\faGithub\ Code}
\quad \href{https://huggingface.co/GAIR/daVinci-Agency}{\raisebox{-.15em}{\includegraphics[height=1em]{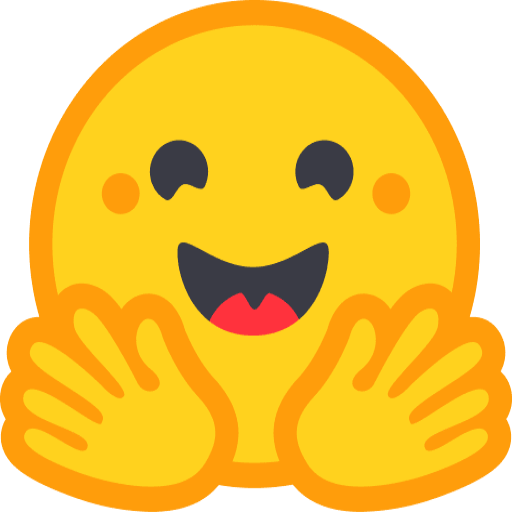}}\ Models}
\quad \href{https://huggingface.co/datasets/GAIR/daVinci-Agency}{\textcolor{violet}\faDatabase\ Datasets}
\end{center}


\vspace{10pt}

\begin{abstract}
While Large Language Models (LLMs) excel at short-term tasks, scaling them to long-horizon agentic workflows remains challenging. The core bottleneck lies in the scarcity of training data that captures authentic long-dependency structures and cross-stage evolutionary dynamics—existing synthesis methods either confine to single-feature scenarios constrained by model distribution, or incur prohibitive human annotation costs, failing to provide scalable, high-quality supervision. 
We address this by reconceptualizing data synthesis through the lens of real-world software evolution. Our key insight: Pull Request (PR) sequences naturally embody the supervision signals for long-horizon learning. They decompose complex objectives into verifiable submission units, maintain functional coherence across iterations, and encode authentic refinement patterns through bug-fix histories. Building on this, we propose \textbf{daVinci-Agency}, which systematically mines structured supervision from \textbf{chain-of-PRs} through three interlocking mechanisms: (1) progressive task decomposition via continuous commits, (2) long-term consistency enforcement through unified functional objectives, and (3) verifiable refinement from authentic bug-fix trajectories. Unlike synthetic trajectories that treat each step independently, daVinci-Agency's PR-grounded structure inherently preserves the causal dependencies and iterative refinements essential for teaching persistent goal-directed behavior and enables natural alignment with project-level, full-cycle task modeling.
The resulting trajectories are substantial—averaging 85k tokens and 116 tool calls—yet remarkably data-efficient: fine-tuning GLM-4.6 on 239 daVinci-Agency samples yields broad improvements across benchmarks, notably achieving a 47\% relative gain on \texttt{Toolathlon}. Beyond benchmark performance, our analysis confirms the model's effective internalization of long-horizon behaviors and unveils training and inference scaling laws specific to extended planning tasks. Our work establishes daVinci-Agency as a scalable paradigm that overcomes the limitations of single-feature synthesis, demonstrating that modeling real-world evolutionary trajectories offers a principled path to unlock the intrinsic long-horizon potential of agents.

\end{abstract}


\begin{figure}[h]
    \centering
    \includegraphics[width=0.82\linewidth]{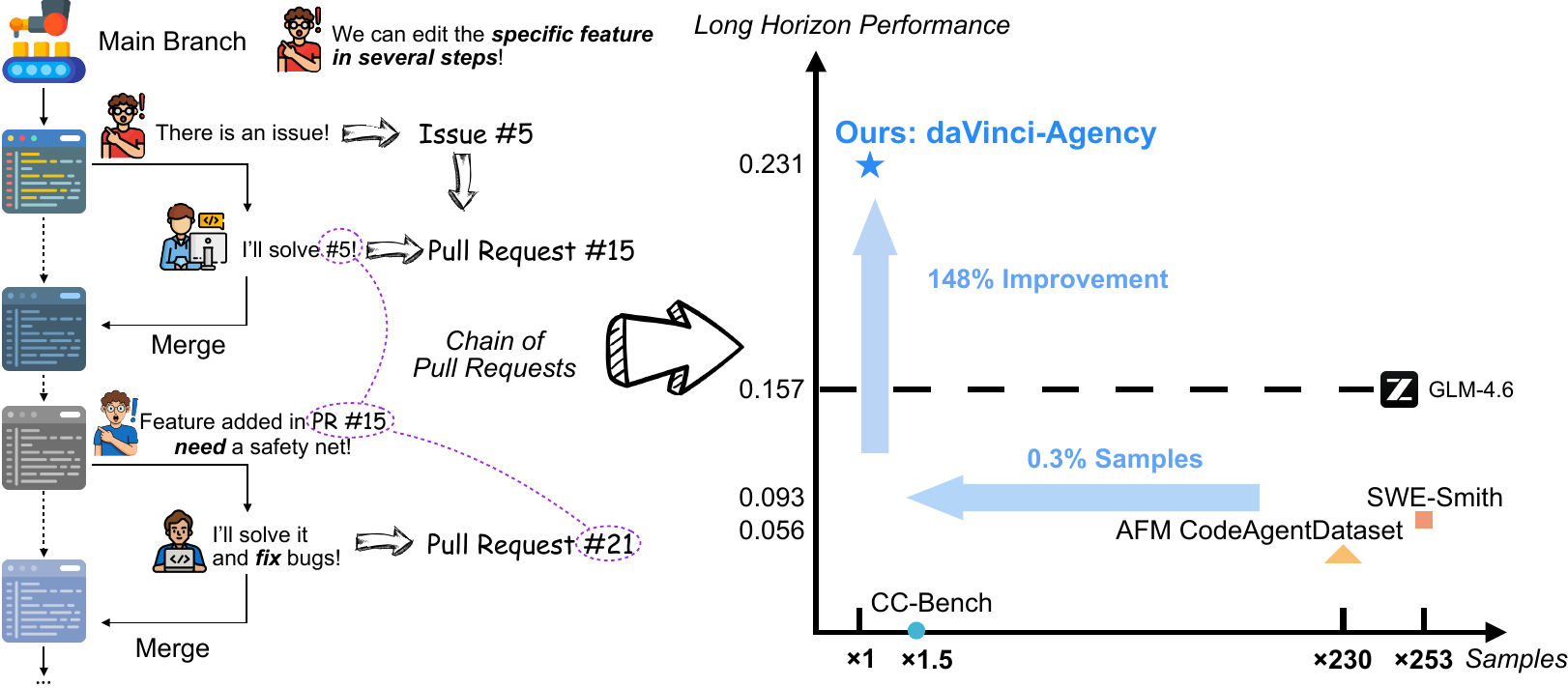}
    \caption{Unlocking long-horizon agency via \textbf{chain of pull requests}. \textbf{Left:} daVinci-Agency extracts supervision for decomposition, consistency, and refinement from the natural evolution of software features. \textbf{Right:} With only 239 training samples, it achieves over 148\% improvement compared to the model trained on 66k samples.}
    \label{fig:teaser}
\end{figure}

\newpage

\pagestyle{fancy}
\lhead{\rightmark}
\renewcommand{\headrulewidth}{0.7pt}
\setlength{\headsep}{5mm}

\clearpage

\newpage

\renewcommand{\thefootnote}{\arabic{footnote}}
\setcounter{footnote}{0}  

\section{Introduction}

With continuous advancements in Large Language Models (LLMs) across code generation, tool calling, and chain-of-thought reasoning~\citep{zeng2025glm,team2025kimi,cai2025nex}, LLM-based agentic system have demonstrated remarkable reliability and usability in short-horizon tasks. Consequently, the research focus is shifting from tasks within a few steps to more challenging long-horizon agentic tasks~\citep{li2025tool,luo2025ultrahorizon,jiang2025adaptation,wu2025innovatorbench}. These tasks typically revolve around a sustained objective, requiring in iterative decision-making, continuous progression, and verifiable delivery over extended interaction spans.

The core challenge of long-horizon tasks lies not merely in extending the length of reasoning, but in the agent's ability to maintain a sense of direction and mitigate cumulative errors over extended cycles. As the task horizon extends, procedural behaviors—such as task decomposition, long-term consistency, and refinement-appear more frequently and become increasingly decisive for success~\citep{guo2025deepseek,xue2024decompose,radhakrishnan2023question,yao2022react}.  Merely relying on single feature tasks is often insufficient for models to acquire these capabilities, as critical difficulties typically emerge only under conditions of cross-stage dependencies and error accumulation, necessitating explicit supervision signals derived from cross-stage evolution. Therefore, long-horizon agent research faces a critical bottleneck: \textit{How to scalably construct long-horizon data where the tasks themselves exhibit sufficiently long dependency structures and cross-stage evolutionary patterns?}

Current approaches to long-horizon data construction generally follow two paradigms: generating interaction trajectories in synthetic environments via distillation or reinforcement learning, or obtaining high-quality supervision through manual annotation~\citep{zhang2025can,wang2023mimicplay,wang2025practitioner,xiao2026mimo}. Existing methods, often limited to single-feature development without cross-stage evolutionary supervision, remain constrained by the generation distribution, which restricts their coverage of realistic failure modes and refinement paths. The latter offers higher-fidelity supervision, yet the processes of collection, annotation, and quality assurance are complex and cost-prohibitive, making them difficult to scale for system training and evaluation.

In this context, GitHub Pull Request (PR) sequences offer a representative candidate data modality. Unlike isolated submissions, software development entails continuous iteration constrained by review and testing. Multiple PRs often align with a single functional goal, progressively advancing delivery through evolving code states, with subsequent PRs frequently addressing prior defects or incorporating feedback-driven refinements. \textbf{Chains of Pull Requests} process naturally embodies the state evolution and external verification signals required for long-horizon interactions, providing a realistic foundation for cross-stage long-cycle task modeling.

Inspired by this, we propose \methodShort, a novel data synthesis paradigm for long-horizon agent learning. Grounded in the continuous PR evolution of real-world software development as Figure~\ref{fig:evolution}, \methodShort automatically organizes multiple PRs centering on a common objective into interdependent task chains. This fosters long-horizon interactions and feedback within evolving code states, thereby extending single-patch generation into a multi-turn submission-feedback-refinement trajectory. Under this setting, the agent must not only achieve phase-wise progression but also maintain overall goal consistency and perform effective error correction under constraints of evolving code states and external feedback. In practice, by chaining up to five PRs, we construct task chains yielding trajectories with an average length of 85k tokens and 116 tool callings, providing scalable supervision for training and evaluating agents' capabilities in long-term planning, consistency, and iterative improvement.

Our experiments demonstrate the effectiveness of \methodShort in long-horizon tasks. Fine-tuning GLM-4.6 on merely 239 such samples yields broad improvements on general benchmarks, notably a 47\% gain on Toolathlon. These results underscore the immense potential of long-horizon synthetic trajectories in unlocking the complex task-solving capabilities of autonomous agents.

Furthermore, our case study reflects from a behavioral level that the fine-tuned model exhibits stronger performance in task decomposition, long-term consistency, and refinement. This enables more stable planning of stage goals, maintenance of global alignment, and effective error correction across multi-round feedback. Simultaneously, by modulating the interaction temporal domain, we explore the scaling properties of \ourDataset, demonstrating that extending longer long-horizon closed-loop training data is an effective path to continuously breaking through the performance ceiling of agents under larger long-horizon reasoning budgets.

Our contributions:
\begin{itemize*}
    \item We propose daVinci-Agency, a paradigm utilizing real-world PR evolution to construct verifiable long-horizon training data. Fine-tuning GLM-4.6 on just 239 daVinci-Agency samples yields broad improvements, notably a 47\% relative gain on Toolathlon.
    \item We demonstrate significant efficiency improvements through real-world long-horizon task modeling. Moreover, case studies focusing on three critical meta-skills confirm that daVinci-Agency enables robust planning and error correction.
    \item We unveil data scaling laws specific to long-horizon tasks, establishing that expanding both the training horizon and inference-time interaction budgets is a critical mechanism for continuously advancing agentic performance.
\end{itemize*}

\section{Related Works}
\paragraph{Agentic Language Model}

The development of Agentic Language Models signifies a fundamental paradigm shift from passive text generation toward autonomous decision-making systems. Early seminal contributions established the groundwork for these capabilities: Toolformer~\citep{schick2023toolformer} demonstrated that language models can learn to invoke external APIs in a self-supervised manner, while ReAct~\citep{yao2022react} synergized reasoning and acting by enabling models to interleave reasoning traces with task-specific actions. Building upon these advancements, research has increasingly focused on foundation models natively optimized for agentic performance. For instance, GLM-4.5~\citep{zeng2025glm} provides a unified framework for reasoning, coding, and agentic tasks, significantly enhancing tool-calling success rates through hybrid reasoning modes. Similarly, Kimi-K2~\citep{team2025kimi} introduces a trillion-parameter Mixture-of-Experts (MoE) architecture specifically tailored for agentic intelligence, featuring native tool-use capabilities and verifiable reward training. Recent studies~\citep{wu2025innovatorbench,tau2bench,taubench,li2025tool} indicate that a central challenge for agentic systems lies in precisely calibrating action trajectories through continuous external feedback. Tool use serves not only as the primary interface with external environments but also as a concentrated manifestation of an agent's adaptation capacity. However, existing models often encounter limitations in complex, multi-turn interactions, such as incoherent tool invocation or inefficient utilization of environmental feedback. To address these bottlenecks, the \methodShort constructs trajectory data with exceptionally high interaction density via long-horizon, multi-turn logic. By leveraging the inherent characteristics of these extended dialogues, it mandates that the agent perform high-frequency tool calling and environmental state interactions throughout the reasoning process. This training paradigm, grounded in complex interaction sequences, significantly heightens the agentic model's sensitivity to tool execution outcomes and its feedback processing capabilities. 

\paragraph{Long-Horizon Agency}

As the capabilities of autonomous agents continue to evolve, both academia and industry have increasingly focused on the demands and challenges associated with solving long-horizon tasks. Following the introduction of SWE-bench~\citep{jimenez2023swe}, subsequent benchmarks such as Toolathlon~\citep{li2025tool}, UltraHorizon~\citep{luo2025ultrahorizon}, and SWE-bench Pro~\citep{deng2025swe} have further intensified the requirements for agentic models, necessitating execution across extended time scales and complex decision dimensions. While recent works like SWE-rebench~\citep{badertdinov2025swe} and SWE-Smith~\citep{yang2025swe} attempt to generate long-horizon data by constructing GitHub-based interactive environments, these methods still face a fundamental limitation where existing data synthesis paradigms struggle to explicitly cultivate skills centered on long-horizon execution, as shown in Figure~\ref{fig:case_study}, specifically task decomposition, long-term consistency, and iterative refinement. Since these methods largely rely on outcome supervision or trajectories generated by teacher-bounded models, they often treat these critical skills as implicit byproducts rather than trainable objectives. Consequently, although the generated trajectories extend environmental execution, they lack the necessary skills as cognitive capabilities required to solve practical engineering problems. To bridge this research gap, we propose the \methodShort framework. By mining the natural evolutionary process of code within Pull Requests, our framework extracts explicit supervision signals for these skills, which are difficult to obtain via manual annotation, thereby constructing high-quality long-horizon task data that not only scales in size but also effectively teaches agents how to execute long-horizon workflows.
\section{Preliminary}
In this section, we formalize the problem of long-horizon agentic software engineering, define the interaction protocol between the agent and the environment, and establish the mathematical notation for the chain-of-PRs used in our data construction.
\begin{figure}[t]
    \centering
    \includegraphics[width=0.88\linewidth]{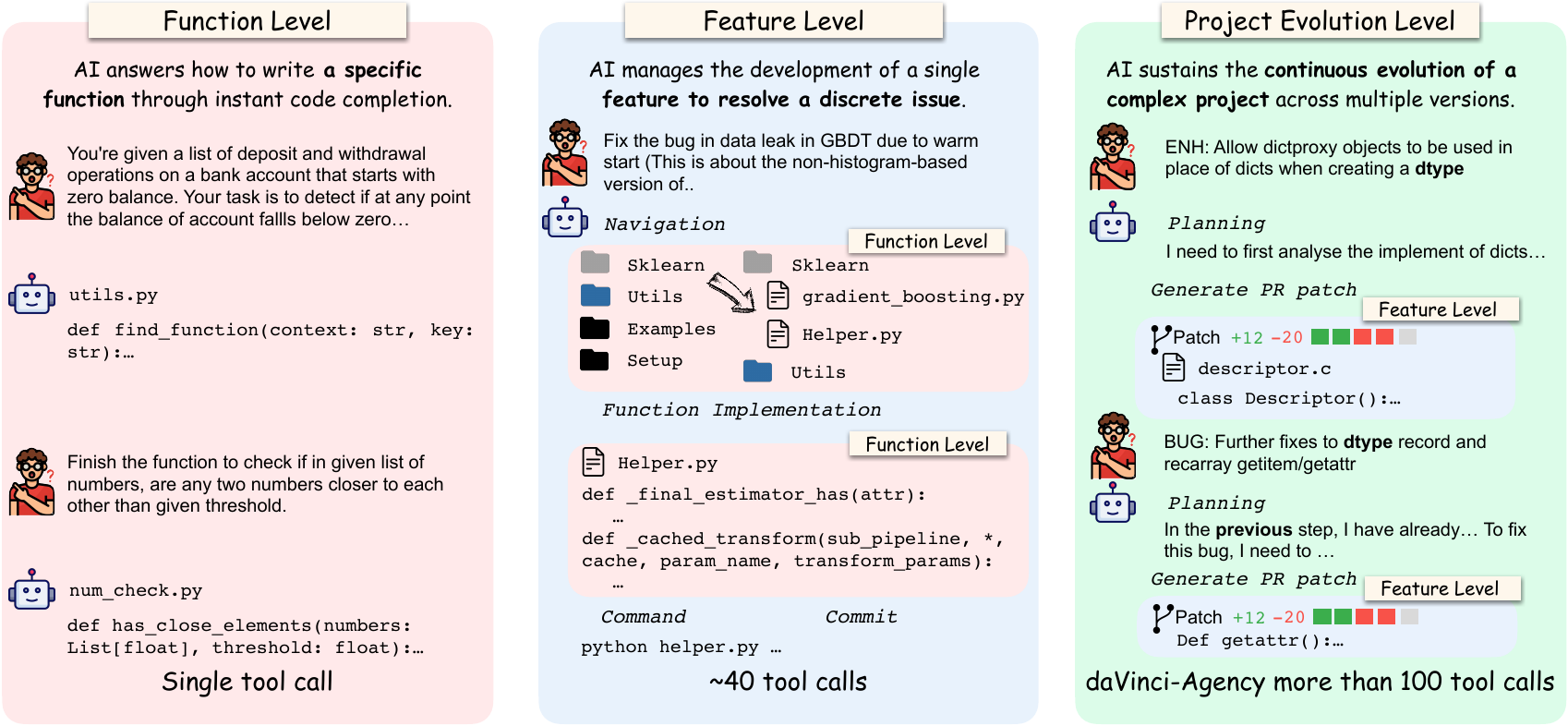}
    \caption{Comparison of scope across software engineering horizons. While function level and feature level focus on isolated algorithms or single-feature resolutions, project evolution level demands that agents handle the continuous evolutionary lifecycle of a project.}
    \label{fig:evolution}
\end{figure}
\subsection{Agent-Environment Interaction Paradigm}
We conceptualize the autonomous software engineering task as a sequential decision-making process modeled as a Markov Decision Process~\citep{puterman1990markov}. The interaction involves an agent $\mathcal{A}$ and a software environment $\mathcal{E}$ via an execution scaffold.

At each time step $j$, the agent observes the current state $o_{j} \in \mathcal{O}$ and executes an action. We categorize the agent's actions into two modalities:
\begin{itemize}
    \item Reasoning Message ($m_j$): Internal cognitive processes where the agent plans, reflects, or analyzes the current context without altering the environment.
    \item Tool Execution ($t_j$): External actions that manipulate the environment (e.g., file editing, running tests, git operations).
\end{itemize}
Formally, a complete interaction trajectory $\tau$ for a specific task is defined as a sequence of these interleaved events~\citep{xiao2025limi,yao2022react}:\begin{equation}\tau = { (o_0, m_0, t_0), (o_1, m_1, t_1), \dots, (o_N, m_N, t_N) }\end{equation}where $N$ represents the horizon length. The goal of the agent is to generate a trajectory that transforms the repository from an initial state $S_{\text{init}}$ to a target state $S_{\text{target}}$ that satisfies a given query $q$.

\subsection{Software Evolution and Chains of Pull Requests}
To capture the continuous nature of software development, we model a repository $\mathcal{R}$ not as a static snapshot, but as an evolving entity driven by PRs.
\paragraph{Pull Request}
Let a single Pull Request be denoted as $pr = (x, \hat{p})$, where $x$ represents the natural language context (including issue descriptions, commit messages, and comments), and $\hat{p}$ represents the ground truth patch (code modification) applied to the repository.
\paragraph{Dependency Topology and PR Chains}
Real-world development is characterized by temporal and semantic dependencies. We define a PR Chain $\mathcal{C}$ as an ordered sequence of $k$ semantically linked PRs:\begin{equation}\mathcal{C} = {pr_1, pr_2, \dots, pr_k} \quad \text{where} \quad pr_{i} \xrightarrow{\text{ref}} pr_{i-1}\end{equation}Here, the relation $\xrightarrow{\text{ref}}$ denotes a dependency where $pr_i$ iterates upon, fixes, or extends the features introduced in previous PRs. Unlike isolated tasks (e.g., standard issue fixing), solving a task within $\mathcal{C}$ requires the agent to maintain long-horizon context and perform state management across the evolving codebase, as the successful implementation of $pr_i$ is contingent upon the correct resolution of $pr_{i-1}$.

\subsection{Task Formulation}
Let $\pi_{\theta}$ denote the agent's policy parameterized by $\theta$. Given a task query $q$ derived from a PR chain $\mathcal{C}$, the agent interacts with the environment to instantiate a reasoning trajectory $\tau$. At each step $j$, the agent samples reasoning and tool execution actions $(m_j, t_j) \sim \pi_{\theta}(\cdot | o_{\le j}, m_{<j}, t_{<j}, \mathcal{C}, q)$ conditioned on the interaction history, culminating in a final submission patch $p$. To guarantee the fidelity of the synthetic data, we employ a rigorous evaluation function defined as $(\text{comment}, s) = E(q, p, \hat{p})$ to filter rollout results based on where the evaluation score $s$ and the ground truth patch $\hat{p}$. This rejection sampling process yields a high-quality dataset $\mathcal{D}_{train} = \{ (\mathcal{C}, q, \tau) \mid s\ge 0.8 \}$, which serves as the basis for policy optimization. Consequently, the training objective is defined as minimizing the negative log-likelihood loss 
\begin{equation}
    \mathcal{L}(\theta) = - \mathbb{E}_{(\mathcal{C}, q, \tau) \sim \mathcal{D}_{train}} \left[ \sum_{t=0}^{T} \log \pi_{\theta}(a_t | \mathcal{C}, q, o_{\le t}, a_{<t}) \right]
\end{equation}
This optimization process crucially extracts supervision for skills from cross-stage evolution in Figure~\ref{fig:evolution}: the sequential query formulation enforces task recomposition, while the rigorous alignment with evolving ground truth states instills long-term consistency and refinement strategies.


\begin{figure}[t]
    \centering
    \includegraphics[width=\linewidth]{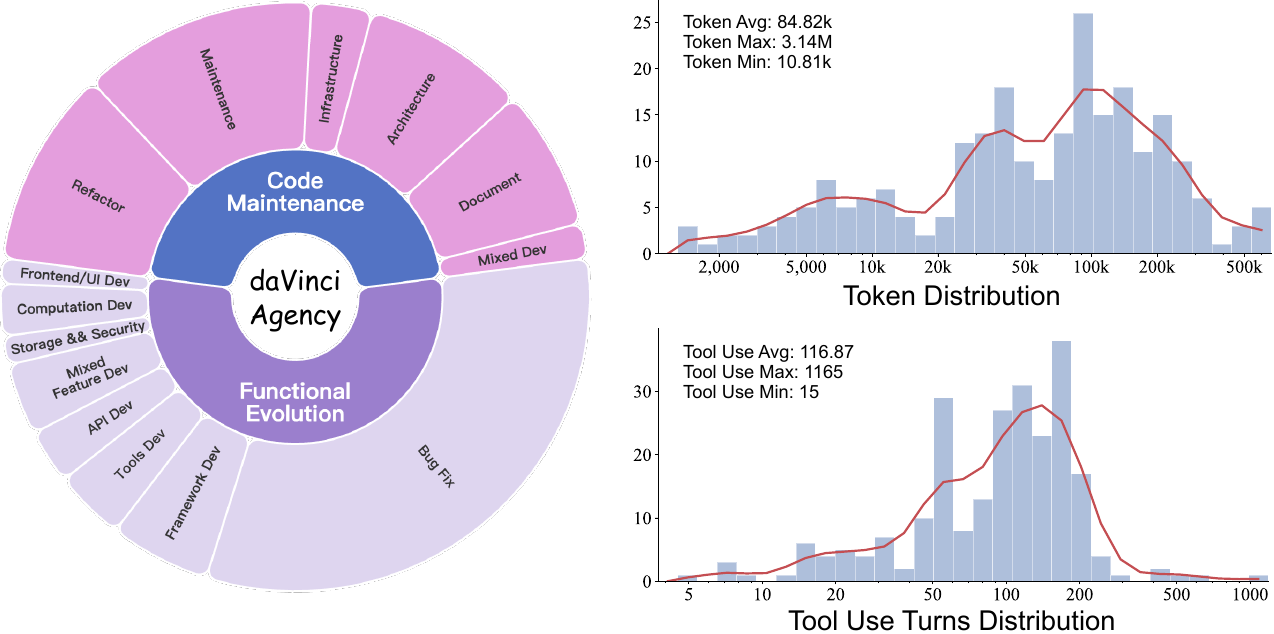}
    \caption{Characteristics of \ourDataset data. Left: Domain coverage across \ourDataset. Right: Distributions of trajectory length and tool utilization illustrating the significant complexity inherent in long-horizon agentic tasks.}
    \label{fig:distribution}
\end{figure}
\section{daVinci-Agency: A Long-Horizon Data Synthesis Paradigm}

\begin{figure}[t]
    \centering
    \includegraphics[width=\linewidth]{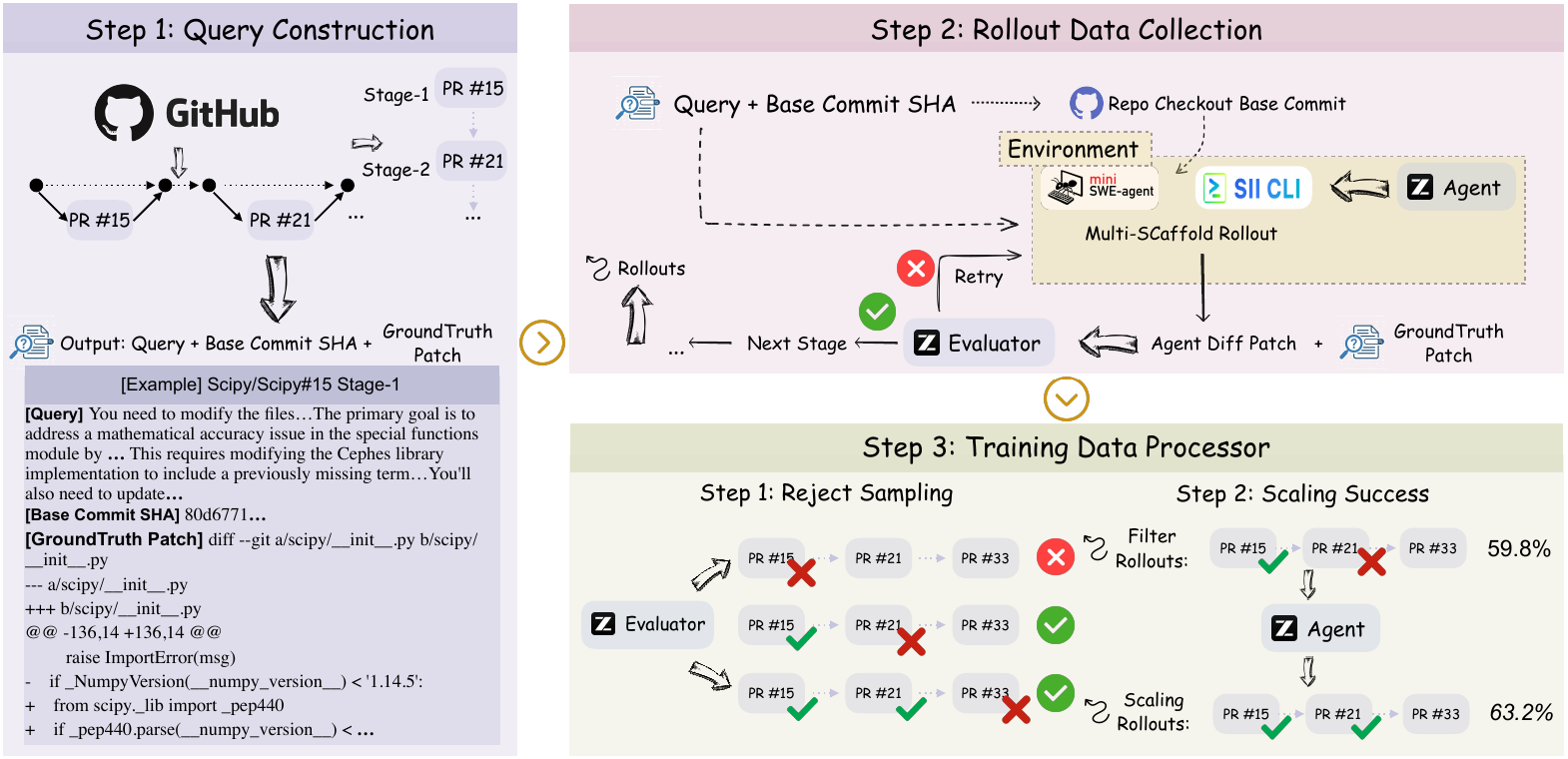}
    \caption{Overview of the \methodShort Data Synthesis Paradigm. The pipeline initiates with query construction, mining PRs with dependency structures from GitHub to form topological task chains, providing reliable state evolution signals as supervision.}
    \label{fig:limi_pipeline}
\end{figure}

The effectiveness of the chain-of-PRs stems from capturing the skills inherent in long-horizon agentic behaviors through continuous feature development evolution. As illustrated in Figure~\ref{fig:limi_pipeline}, this section details our data sourcing and curation pipeline.

\subsection{Data Resource}

We expertly curated nine representative GitHub repositories as our primary data~\citep{zeng2026davinci} sources shown in Table~\ref{tab:repos}, adhering to a rigorous expert-led selection protocol to ensure high data integrity and diversity. Specifically, we conducted qualitative assessments based on three pillars: (1) Scale and Maturity, selecting repositories with over 7,000 effective PRs to ensure a rich history of complex inter-PR dependencies; (2) Community Interactivity, targeting environments with high-frequency code reviews to provide rich natural language reasoning signals; and (3) Linguistic Diversity, where repositories were hand-picked to span diverse technology stacks including but not limited to Python, Java, and C. As illustrated in Figure~\ref{fig:distribution}, this meticulous, human-in-the-loop selection ensures that our long-horizon data comprehensively reflects complex, heterogeneous real-world development scenarios.

\subsection{Data Curation Pipeline}

To construct long-horizon scenarios based on the aforementioned GitHub data sources, we structure our data curation pipeline into two distinct stages: PR chain construction and query formulation. This process formalizes the query generation function as $q=f(x,\hat{p},\mathcal{R})$, which serves as the initial input for the subsequent agentic rollout.
\paragraph{PR-Chain Combination}

We construct PR development chains by leveraging metadata from the GitHub API, utilizing explicit citations within commit messages and review comments to instantiate a dependency topology $\mathcal{C} = \{pr_1, \dots, pr_k\}$ that enforces the strictly ordered relation $pr_{i} \xrightarrow{\text{ref}} pr_{i-1}$. As illustrated in Figure~\ref{fig:limi_pipeline}, this topology captures the non-contiguous nature of real-world iteration (e.g., PR \#21 iterating upon PR \#15). By faithfully reconstructing this authentic process of feature evolution, our method empowers models to transcend isolated problem-solving, enabling them to master step-by-step state management and long-term consistency within a continuously evolving codebase.

\paragraph{Query Construction}

Building upon the constructed PR chains, we formalize the query generation process as $q=f(x, \hat{p}, \mathcal{R})$, taking the natural language context $x$ and the ground truth patch $\hat{p}$ from the PR tuple. Specifically, we leverage an LLM to synthesize a corresponding sub-query for each PR within the chain. These sub-queries are designed to elucidate the core problem addressed in the current step and the underlying reasoning chain, while also explaining the roles of relevant symbols and variables. Crucially, we intentionally withhold specific implementation details within these sub-queries. This constraint compels the agent to fully exercise its code navigation and localization capabilities during the rollout process. Furthermore, to equip the agent with a macroscopic perspective, we augment the initial query with a global overview of the entire PR chain, ensuring that the agent establishes a comprehensive understanding of the long-horizon task's global objectives and structural decomposition from the outset.

\subsection{Rollout Data Collection for Training Dataset}

\paragraph{Environment for Rollout}
To faithfully simulate real-world incremental software development workflows, we construct a sampling environment designed for sequential stage-by-stage execution. Within this framework, the agent is required to address tasks sequentially along the PR chain. To enforce continuity, we devise a state transition mechanism based on file modification propagation, where the agent's code changes from the previous stage are explicitly carried over to the next. Formally, let $B_t$ denote the base codebase (commit) for stage $t$, and $\Delta\tau_{t-1}$ represent the cumulative code patch generated by the agent's trajectory in the previous stage. The agent's rollout for the current stage $\tau_t$ is sampled conditioned on this evolved state:
\begin{equation}\
\tau_t \sim \pi_\theta \left( \cdot \mid S_{\text{init}}^{(t)}, q_t \right), \quad \text{where} \quad S_{\text{init}}^{(t)} = B_t \oplus \Delta\tau_{t-1}
\end{equation}
Here, $\oplus$ denotes the operation of applying the previous patch onto the new base branch, and $q_t$ is the task query. This recursive dependency ensures that subsequent tasks are strictly built upon the agent's own evolved codebase state, thereby compelling the model to address the issues of long-term consistency and refinement following error accumulation in long-horizon interactions under supervision.

\paragraph{Evaluator for Rejection Sampling}

Rejection sampling is pivotal for ensuring data quality and downstream performance as shown in Table~\ref{tab:reject_sampling}. We employ GLM-4.6~\cite{zeng2025glm} to evaluate the semantic alignment between the generated patch $p$ and the ground truth $\hat{p}$. With a strict threshold of $s \ge 0.8$, the evaluator provides textual feedback for up to three refinement iterations. Trajectories failing to meet this standard are discarded, ensuring that only samples with high semantic fidelity are retained.

\paragraph{Rollout Scaffold}
We deploy GLM-4.6 across two distinct scaffolds—SII-CLI~\citep{sii-cli-2025} and mini-swe-agent~\citep{yang2024swe}—to capture diverse agentic behaviors. The framework logs complete interaction trajectories $\tau_i = \{(o_{i,j}, m_{i,j}, t_{i,j})\}_{j=0}^{N_i}$, encompassing observations, reasoning, and tool execution. These scales underscore the complexity of real-world software evolution and the critical need for robust long-context coherence.

\section{Experiments}
 \begin{table}[t]
\centering
\resizebox{\textwidth}{!}{
\begin{tabular}{lcccccccc}
\toprule
\textbf{Training Data}                          & \textbf{Samples} & \textbf{SWE-bench} & \textbf{Toolathlon} & \textbf{DS-1000} & \textbf{\makecell{$\tau^2$-bench\\-retail}} & \textbf{\makecell{$\tau^2$-bench\\-airline}} & \textbf{SciCode-MP} & \textbf{AVG} \\
\midrule
\textbf{GLM-4.6}      & - & 0.608 & 0.157 & 0.522 & 0.675 & \textbf{0.620} & 0.062 & 0.441 \\
\textbf{SWE-Smith}              & 66,000           & 0.404              & 0.093               & 0.470            & 0.586                      & 0.565                       & 0.123               & 0.373        \\
\textbf{CC-Bench}             & 260              & 0.618              & 0.000               & 0.526            & 0.697                      &0.605                       & 0.169               & 0.436        \\
\textbf{CodeAgent} & 59,939           & 0.184              & 0.056               & 0.308            & 0.564                      & 0.560                       & 0.015               & 0.281        \\
\midrule
\textbf{$\text{\methodShort}_\text{SinglePR}$}                 & 2786             & 0.166              & 0.120               & 0.396            & 0.590                      & 0.465                       & 0.154               & 0.315        \\
\textbf{$\text{\methodShort}_\text{TemporalChain}$}           & 600              & 0.616              & 0.213               & \textbf{0.541}            & 0.472                      & 0.525                       & \textbf{0.185}               & 0.425        \\
\cellcolor{coolblue3}\textbf{\methodShort}                           & \cellcolor{coolblue3}239              & \cellcolor{coolblue3}\textbf{0.632}              & \cellcolor{coolblue3}\textbf{0.231}               & \cellcolor{coolblue3}0.526            & \cellcolor{coolblue3}\textbf{0.707}                      & \cellcolor{coolblue3}0.597                       & \cellcolor{coolblue3}0.154               & \cellcolor{coolblue3}\textbf{0.475}       \\
\bottomrule

\end{tabular}

}
\caption{Performance comparison across diverse long-horizon datasets on GLM-4.6. Despite being trained on only 239 samples, \methodShort outperforms baselines trained on datasets that are orders of magnitude larger and incorporate executable feedback within their trajectories. SciCode-MP denotes SciCode main problem.}
\label{tab:effiency}
\end{table}

\subsection{Experiment Setups}
\paragraph{Baselines}

To evaluate the performance of our fine-tuned models, we benchmarked them against a diverse set of state-of-the-art open-source models, including GLM-4.6~\citep{zeng2025glm}, Kimi-K2-thinking~\citep{team2025kimi}, DeepSeek-v3.2~\citep{liu2025deepseek}, Qwen3-235B-A22B-Instruct~\citep{yang2025qwen3}, and Qwen3-30B-A3B-Thinking-2507~\citep{yang2025qwen3}. These baselines span a wide spectrum of architectural designs and parameter scales, ensuring a rigorous and comprehensive performance comparison.
\paragraph{Model Training and Variants}

To comprehensively evaluate the effectiveness of \ourDataset, we selected four representative high-performance models, including GLM-4.6~\citep{zeng2025glm}, Qwen3-30B-A3B~\citep{yang2025qwen3}, Qwen3-32B~\citep{yang2025qwen3}, and Qwen3-8B for supervised fine-tuning cross different model architectures and scales. 
Regarding implementation, we performed full-parameter fine-tuning on all aforementioned models utilizing the slime framework~\citep{slime_github}. To ensure the fairness and rigor of our comparative analysis, we strictly controlled for hyperparameter consistency.

To evaluate the effectiveness of \ourDataset, we categorize the comparison sets into external \textit{baseline datasets} and internal \textit{dataset variants}. For baseline datasets, we selected three representative datasets in the field of agentic model training as baselines for comparison:
\begin{itemize*}
    \item SWE-Smith~\citep{yang2025swe}: Agentic trajectories collected using the SWE-agent scaffold with Claude-3.7-Sonnet as the rollout model.
    \item CC-Bench~\citep{zeng2025glm}: Agentic trajectories collected using the Claude Code scaffold with sota models such as Claude-4-Sonnet and Kimi-K2 as rollout models.
    \item AFM CodeAgentDataset (denoted as CodeAgent in the table)~\citep{li2025chainofagentsendtoendagentfoundation}: Code Q\&A agentic trajectories collected using models like Claude as the rollout model.
\end{itemize*}
For dataset variants, we introduce two variants to investigate specific contributions of design components within \methodShort:
\begin{itemize*}
    \item $\text{\methodShort}_\text{SinglePR}$ restricts tasks to isolated pull requests, serving as a baseline to demonstrate the necessity of the multi-PR paradigm for modeling long-horizon dependencies.
    \item $\text{\methodShort}_\text{TemporalChain}$ constructs sequences by concatenating chronologically adjacent PRs rather than leveraging semantic reference relationships.
\end{itemize*}

\paragraph{Benchmarks}

To comprehensively evaluate the model's capabilities in long-horizon agentic tasks, we curated a suite of five distinct benchmarks. Specifically, we utilize SWE-bench Verified~\citep{jimenez2023swe} to assess feature-level software engineering proficiency. Toolathlon~\citep{li2025tool} and $\tau^2$-Bench~\citep{tau2bench} are employed to evaluate tool utilization and interactive capabilities, while DS-1000~\citep{ds1000} and SciCode~\citep{SciCode} serve as benchmarks for file-level coding performance.

Regarding the evaluation protocol, for SWE-bench, we adopt SWE-Agent~\citep{yang2024swe} as the execution scaffold. For Toolathlon, DS-1000, and SciCode, we strictly adhere to the experimental settings described in their respective original publications. In the case of $\tau^2$-Bench, we employ GLM-4.6 to simulate the user agent across all model evaluations and evaluate both the retail and airline domains.

\begin{table}[t]
\centering
\resizebox{\textwidth}{!}{
\begin{tabular}{lccccccc}
\toprule
\textbf{Model}   & \textbf{SWE-bench} & \textbf{Toolathlon} & \textbf{DS-1000} & \textbf{\makecell{$\tau^2$-bench\\-retail}} & \textbf{\makecell{$\tau^2$-bench\\-airline}} & \textbf{SciCode-MP} & \textbf{AVG} \\
\midrule
\textbf{DeepSeek-v3.2 }           & {0.456}   & \textbf{0.250}                   & 0.320            & 0.640                      & 0.530                       & 0.000               & 0.366        \\
\textbf{Kimi-K2-Thinking}         & 0.318        &    0.213                     &   0.521               &    0.702                        &     0.610                        &   0.062                  &  0.404            \\
\textbf{Qwen3-235B} & { 0.504}   & 0.046                   & 0.258            & 0.610                      & 0.435                       & 0.000               & 0.309        \\
\textbf{Qwen3-4B}      & 0.098 & 0.009 & 0.256 & 0.338 & 0.285 & 0.000 & 0.164 \\
\cellcolor{coolblue3}\;\;+ \ourDataset            & \cellcolor{coolblue3}0.116 & \cellcolor{coolblue3}0.019 & \cellcolor{coolblue3}0.261 & \cellcolor{coolblue3}0.360 & \cellcolor{coolblue3}0.255 & \cellcolor{coolblue3}0.000 & \cellcolor{coolblue3}0.168 \\
\textbf{Qwen3-32B}     & 0.298 & 0.019 & 0.398 & 0.536 & 0.415 & 0.015 & 0.280 \\
\cellcolor{coolblue3}\;\;+ \ourDataset            & \cellcolor{coolblue3}0.310 & \cellcolor{coolblue3}0.037 & \cellcolor{coolblue3}0.410 & \cellcolor{coolblue3}0.526 & \cellcolor{coolblue3}0.435 & \cellcolor{coolblue3}0.031 & \cellcolor{coolblue3}0.292 \\
\textbf{Qwen3-30B-A3B} & 0.242 & 0.037 & 0.370 & 0.579 & 0.510 & 0.031 & 0.295 \\
\cellcolor{coolblue3}\;\;+ \ourDataset            & \cellcolor{coolblue3}0.262 & \cellcolor{coolblue3}0.046 & \cellcolor{coolblue3}0.343 & \cellcolor{coolblue3}0.608 & \cellcolor{coolblue3}0.540 & \cellcolor{coolblue3}0.046 & \cellcolor{coolblue3}0.307 \\
\textbf{GLM-4.6}       & 0.608 & 0.157 & 0.522 & 0.675 & \textbf{0.620} & 0.062 & 0.441 \\
\cellcolor{coolblue3} \;\;+ \ourDataset          & \cellcolor{coolblue3}\textbf{0.632} & \cellcolor{coolblue3}0.231 & \cellcolor{coolblue3}\textbf{0.526} & \cellcolor{coolblue3}\textbf{0.707} & \cellcolor{coolblue3}0.597 & \cellcolor{coolblue3}\textbf{0.154} & \cellcolor{coolblue3}\textbf{0.475} \\
\bottomrule
\end{tabular}

}
\caption{Comprehensive comparison against baselines and cross-model fine-tuning analysis. The fine-tuning dataset \ourDataset is synthesized using GLM-4.6 as the rollout model.}

\label{tab:main_result}
\end{table}

\subsection{Main Results}
We evaluated models trained on three external agentic datasets alongside ablation subsets of \ourDataset. The results are shown in Table~\ref{tab:effiency}.
\paragraph{Performance against Agent Datasets}
The results demonstrate that despite its smaller sample size, \ourDataset consistently outperforms the aforementioned datasets across all benchmarks. Notably, due to the lack of cross-stage evolution, SWE-Smith struggles to achieve comparable gains despite its large data scale. In contrast, by introducing cross-stage evolution through the modeling of real-world development tasks, \ourDataset uniquely achieved significant gains on Toolathlon while maintaining robustness on SWE-bench. This underscores that mere environment-based distillation captures only surface-level behaviors, whereas \ourDataset extracts structured supervision from PR evolution to effectively equip models with essential long-horizon agency, demonstrating its core advantage in building robust agents.

\paragraph{Impact of Data Curation}

To validate the critical role of the chain-of-PRs, we conducted a rigorous ablation study. Results show that $\text{\methodShort}_\text{SinglePR}$, lacking cross-stage evolution within PRs mechanism, fails to provide supervision for task decomposition, long-term consistency, and refinement via fix PRs, resulting in decreased performance. Meanwhile, the chronologically ordered $\text{\methodShort}_\text{TemporalChain}$, despite lacking explicit semantic links, yields limited long-horizon benefits by implicitly reinforcing task decomposition through repository-level context. In contrast, our \methodShort achieves superior performance with higher data efficiency. By preserving the precise logical dependencies of authentic code development, \methodShort provides comprehensive supervision from real evolution of PRs, confirming that intrinsic evolutionary logic—not just sequence length—is the determinant for mastering complex agentic long-horizon tasks.

\subsection{Model Performance Comparison}
\begin{wraptable}{r}{0.4\textwidth}
    \centering
    \resizebox{0.4\textwidth}{!}{
\begin{tabular}{lc}
\toprule
\textbf{Model}                    & \textbf{\makecell{AgencyBench}} \\
\midrule
\textbf{GLM-4.6}                  & 11.9                      \\
\textbf{DeepSeek-v3.2}            & 11.6                      \\
\textbf{Kimi-K2-Thinking}         & 11.8                      \\
\textbf{Qwen3-235B-A22B-Instruct} & 4.6                       \\
\textbf{GLM-4.6-\ourDataset}                     & \textbf{15.9}                     \\
\bottomrule
\end{tabular}
}
    \caption{Performance on multi-turn, long-horizon tasks in the AgencyBench Code.}
    \label{tab:agency}
\end{wraptable}
Experimental results in Table~\ref{tab:main_result} demonstrate that training on \ourDataset confers a substantial advantage in long-horizon agentic tasks, achieving a superior overall average score of 0.475. Specifically, our model secures a clear lead in software engineering on SWE-bench, significantly outperforming competitors like GLM-4.6 and Qwen3-235B. This superiority is particularly evident in the multi-turn long-horizon AgencyBench Code~\citep{li2026agencybench} shown in Table~\ref{tab:agency}, where our model achieves a remarkable score of 15.9. Furthermore, it exhibits remarkable proficiency in tool utilization and cross-domain reliability, surpassing Kimi-K2-Thinking on both Toolathlon and $\tau^2$-bench, where it demonstrates the highest stability among tested cohorts.

Even more remarkably, the \methodShort training trajectories are generated exclusively by GLM-4.6 itself, effectively aligning the GLM-4.6 fine-tuning process with the principles of on-policy self-distillation. This distinction is pivotal: it indicates that the substantial performance gains are not derived from knowledge distillation via a superior teacher model, but are attributable solely to the model's self-exploration and optimization within our meticulously curated multi-PR long-horizon tasks. These findings underscore the efficacy of the \methodShort paradigm, demonstrating that unlocking a model's intrinsic potential through high-quality data structures is sufficient to equip agents with the essential skills required to long-horizon agentic tasks.

\subsection{Model Generalization Analysis}

To verify the generalizability of \ourDataset across different model architectures and scales, we conduct fine-tuning experiments on the Qwen3 series, covering both Mixture-of-Experts (MoE) and dense architectures at various parameter counts. The results are reported in Table~\ref{tab:main_result}.

In terms of architectural adaptability, our method demonstrates significant effectiveness on MoE models. \ourDataset proves highly effective on MoE models, raising Qwen3-30B-A3B’s average score to 0.307, with SWE-
bench performance increasing from 0.242 to 0.262. This indicates that our data paradigm successfully transfers complex reasoning patterns to dynamic sparse architectures without compromising their inherent efficiency.

Regarding model scaling, we observe consistent improvements across dense models of varying sizes. For the Qwen3-32B model, our method improves the average score to 0.292, with specific gains in rigorous tasks like SciCode-MP and Toolathlon. Even for the smaller-scale Qwen3-8B, where reasoning capacity is typically constrained, our method still yields a positive uplift in overall performance and coding benchmarks. These results collectively validate that \ourDataset provides a model-agnostic enhancement signal, capable of unlocking agentic potential across a wide spectrum of foundation models.



\begin{figure}[t]
    \centering
    \includegraphics[width=\linewidth]{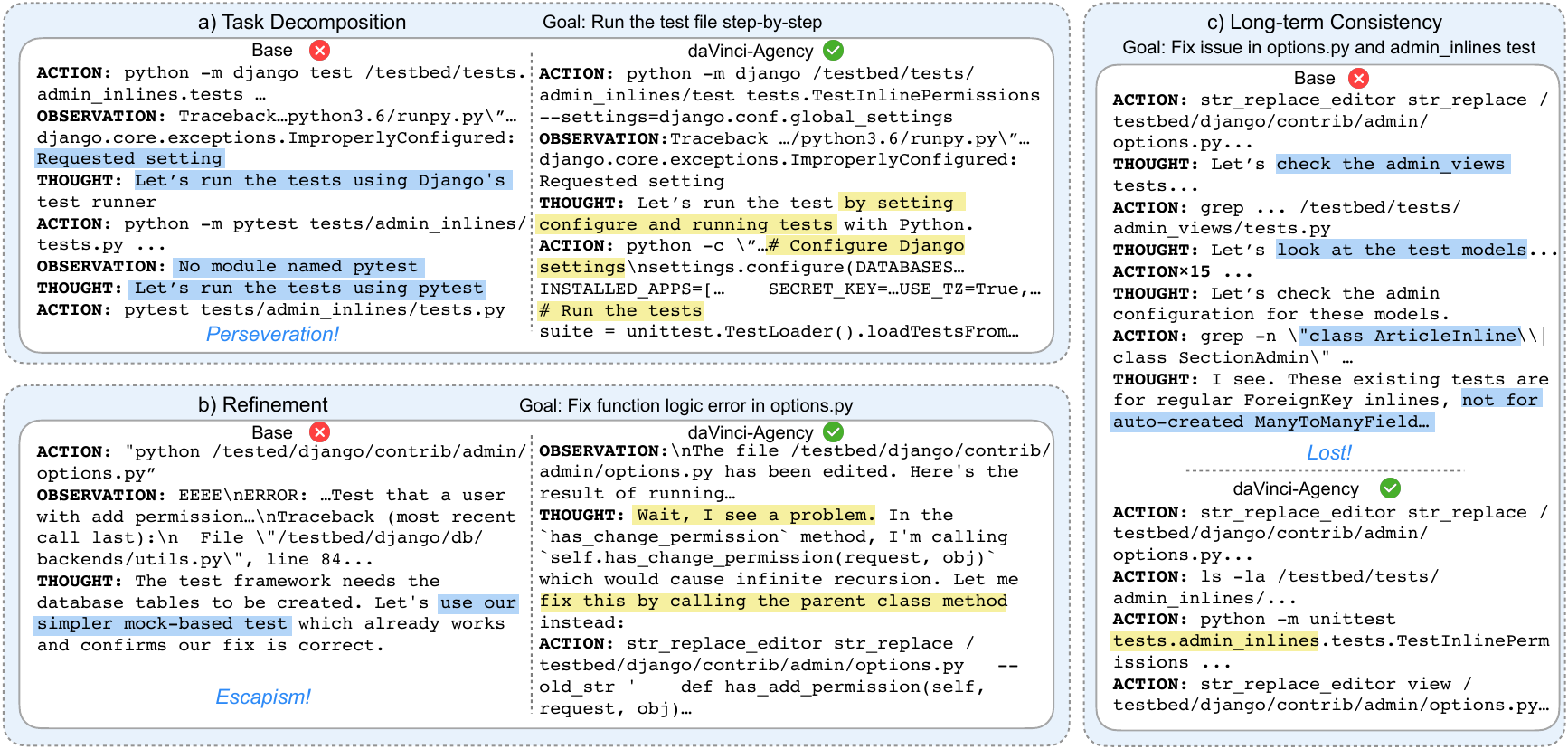}
    \caption{Comparison of behavioral trajectories between the \methodShort and baseline models on the real case in SWE-bench. The \methodShort model demonstrates exceptional long-horizon agency, while the baseline model exhibits a lack of planning, goal drift, and escapism when encountering errors.}
    \label{fig:case_study}
\end{figure}

\section{Long Horizon Analysis}
\subsection{Meta-Skills Performance Comparison}
\begin{figure}[t]
    \centering
    \includegraphics[width=0.9\linewidth]{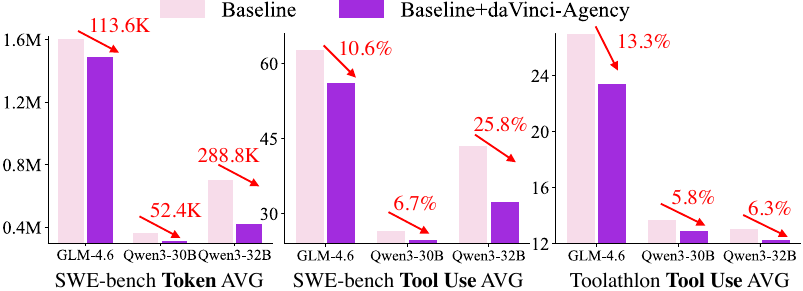}
    \caption{Token and tool use efficiency analysis on long-horizon tasks. Models fine-tuned on \ourDataset achieve higher success rates with significantly reduced token and tool consumption. This superior efficiency stems from the internalization of \methodShort's inherent skills which enable the agent to prune redundant actions and navigate long-horizon tasks with precise planning. Qwen3-30B-A3B denotes as Qwen3-30B.}
    \label{fig:token_effi}
\end{figure}
To visualize the acquisition of long-horizon agency, we analyze the execution trajectories of both GLM-4.6 and GLM-4.6-\ourDataset on SWE-bench Django Issue \#11149 in Figure~\ref{fig:case_study}, a complex permission bug necessitating sustained attentional focus. Despite possessing general coding proficiency, the baseline model exhibits characteristic failures typical of long-horizon agency: it suffers from goal drift, wandering through irrelevant modules like \texttt{admin\_views} for multiple turns. Simultaneously, it displays escapism by disregarding environment configuration errors to rely on simplistic, non-representative scripts. These behaviors highlight a critical deficiency: Meta-skills should be acquired through supervision signals derived from dynamic cognitive behaviors that existing models—constrained by the teacher bound inherent in standard environment-based fine-tuning or distillation—fail to exhibit.

In stark contrast, the \methodShort demonstrates the effective internalization of these implicit skills. It exhibits structured task decomposition by deconstructing the validation process—first establishing configuration and then executing a sequence of test cases from general to specific—rather than relying on a monolithic test suite. Most notably, the model displays an emergent ``aha moment" during the refinement phase: amidst code modification, it pauses to reflect a logic error prior to execution—“Wait, I see a problem... I'm calling self... which would cause infinite recursion”—and proactively implements a fix using \texttt{super()}. This qualitative leap confirms that \methodShort successfully extracts structured supervision from the evolutionary history of software development, unlocking long-horizon agency that standard distillation methods fail to capture.
\subsection{Efficiency Gains in Token and Tool Usage}

As illustrated in Figure~\ref{fig:token_effi}, models trained on \ourDataset demonstrate a remarkable improvement in execution efficiency compared to the baseline. On the SWE-bench benchmark, the average token consumption for GLM-4.6 decreases by 113.6K, while Qwen3-32B sees a substantial reduction of 288.8K tokens. A similar trend is observed in tool usage, with reductions of up to 25.8\% and 13.3\% on SWE-bench and Toolathlon.

We argue that this efficiency is not merely a cost-saving measure but a direct indicator of enhanced intelligence per token. In long-horizon agentic tasks, the context window represents a strictly constrained resource. Excessive verbosity, redundant reasoning steps, or ineffective tool loops dilute the context, heightening the risk of the ``lost-in-the-middle" phenomenon. By leveraging state evolution signals as supervision for long-horizon skills, \methodShort effectively compresses the reasoning trajectory, ensuring that every generated token carries higher information density. This efficiency allows the agent to preserve valuable context space for maintaining long-term dependencies, thereby enabling superior performance on complex, extended-horizon tasks where context management is paramount.
\subsection{Long Horizon Data Scaling}

\begin{figure}[t]
    \centering
    \includegraphics[width=\linewidth]{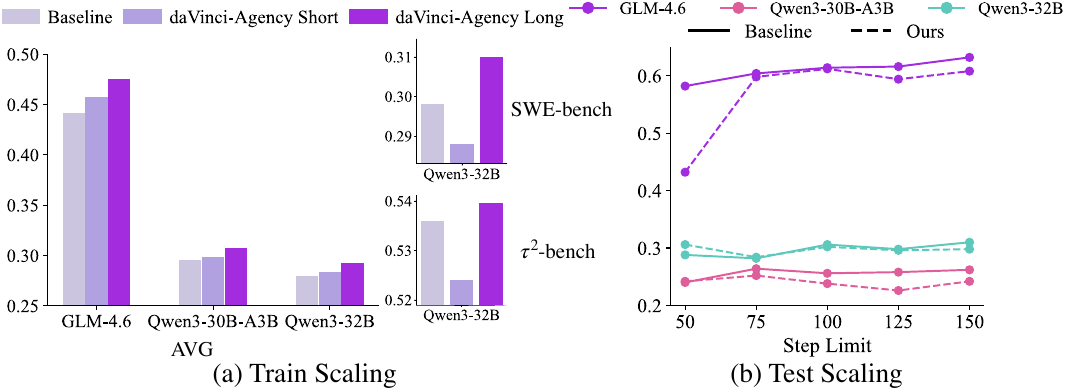}
    \caption{Performance scaling analysis across training horizons and inference budgets. \textbf{Left:} Training Horizon Scaling. Extending training trajectory horizon by scaling PR chains from \ourDataset-Short's 59.39K to \ourDataset-Long's 84.82K tokens significantly enhances model performance across long-horizon benchmarks. \textbf{Right:} Inference-time  Budget Scaling on SWE-bench. A comparison of performance gains under varying inference budgets, demonstrating the model's superior efficiency in leveraging additional test-time compute.}
    \label{fig:train_scaling}
\end{figure}

To investigate the optimal approach for maximizing the efficacy of long-horizon agentic queries constructed via \methodShort, we conducted a comprehensive evaluation of the scaling success strategy, increasing the length of successful trajectories by completing additional pull requests —as illustrated in Figure~\ref{fig:train_scaling}. \ourDataset-Short exhibits an average length of 59.39K tokens per sample. By implementing the scaling success strategy—which extends successful interaction trajectories based on \ourDataset-Short—the resulting \ourDataset-Long achieves a substantially higher average sample length of 
84.82K tokens. This increment demonstrates the effective expansion of sequential data through the completion of long-horizon task chains. Our empirical observations indicate that by completing chains of PRs that were previously left unfinished in the training data, the model’s overall performance is significantly enhanced. Most notably, on the rigorous SWE-bench and $\tau^2$-bench benchmarks, we observed relative performance gains of up to 8\%, validating a direct correlation between extended trajectory length and model robustness.


Even more significantly, the results for Qwen3-32B on SWE-bench highlight a critical phenomenon: this scaling strategy effectively unlocks the latent potential of training data that previously failed to yield performance improvements. This substantial performance leap provides compelling evidence for the pivotal role of long-horizon sequential structures in eliciting latent agentic capabilities. Our experimental results confirm that interaction trajectory length serves as a highly effective scaling dimension for enhancing problem-solving proficiency, thereby establishing a new and promising direction for the data synthesis and training of future long-horizon agents.

Furthermore, to deeply investigate the advantages of models trained on \ourDataset in long-horizon tasks, we conducted a sensitivity analysis on tool call limits across the SWE-bench and $\tau^2$-Bench benchmarks, aiming to assess the model's performance potential under more ample inference budgets. As illustrated in Figure~\ref{fig:train_scaling}, on SWE-bench, the performance gap between the model fine-tuned on \ourDataset and the baselines significantly widens as the allowed tool call limit increases. This phenomenon robustly proves that the long-horizon data paradigm constructed by \ourDataset can significantly enhance model stability in long-horizon contexts. It effectively mitigates the common problem of error accumulation in long sequences when handling complex tasks requiring multi-step reasoning, thereby maintaining a continuous and stable problem-solving capability.

\section{Conclusion}
In this work, we introduce daVinci-Agency to address the critical scarcity of training data that captures authentic long-dependency structures and cross-stage evolutionary dynamics. By mining the natural evolutionary process of real-world software, \methodShort reconstructs semantically dependent PR chains, transforming isolated coding tasks into coherent, multi-stage workflows. This approach provides a scalable, verifiable paradigm for training agents without relying on prohibitively expensive human annotation or teacher-bounded distillation.

This structure compels agents to internalize critical long-horizon behaviors such as task decomposition, long-term consistency, and iterative refinement—that standard distillation methods fail to capture. Empirical results demonstrate remarkable data efficiency; fine-tuning on just 239 \ourDataset samples significantly outperforms baselines trained on much larger datasets, achieving substantial gains on benchmarks like SWE-bench and Toolathlon. This validates that internalizing these skills through high-quality structural data is the decisive factor for mastering complex engineering workflows.

Currently, constrained by the success rates, our implementation connects a maximum of 5 PRs to construct task chains. However, our findings on training horizon scaling suggest that further extending these chains is a viable path for continuous improvement. Future work will focus on overcoming current success rate bottlenecks to construct even longer evolutionary chains, further pushing the boundaries of long-horizon agents.

\newpage
\bibliographystyle{acl_natbib}
\bibliography{main}
\clearpage
\appendix

\section{Repositories in \ourDataset}
\begin{table}[ht]
\centering
\begin{tabular}{cccc}
\toprule
\textbf{Repo ID} & \textbf{Repo Name}    & \textbf{PR Num} & \textbf{Query Num} \\
\midrule
313853293        & cloudquery/cloudquery & 9705            & 490                \\
62117812         & apache/pulsar         & 9803            & 228                \\
1460385          & scipy/scipy           & 9511            & 238                \\
523043277        & astral-sh/ruff        & 8115            & 180                \\
139199684        & PrefectHQ/prefect     & 8433            & 215                \\
908607           & numpy/numpy           & 9355            & 354                \\
680120071        & microsoft/autogen    & 1467            & 18                 \\
330914717        & ivy-llc/ivy          & 4160            & 9                  \\
543276238        & tursodatabase/libsql & 1261            & 30                 \\
\midrule
SUM                 &                       & 61810           & 1762              \\
                 \bottomrule
\end{tabular}

\caption{List of the source repositories used in \ourDataset. The table lists the repository names, the total number of Pull Requests processed (PR Num), and the final number of constructed task queries (Query Num)}

\label{tab:repos}
\end{table}
Table~\ref{tab:repos} presents the detailed statistical breakdown of the nine source repositories selected for \ourDataset, as discussed in the main methodology. These repositories were chosen to ensure a comprehensive coverage of diverse technology stacks and development scales.

As detailed in Table~\ref{tab:repos}, our selection spans a wide spectrum of programming paradigms and domains:
\begin{itemize}
    \item Scientific Computing \& Python Ecosystem: Repositories such as numpy/numpy and scipy/scipy provide rich data on complex mathematical logic and algorithm optimization.
    \item Distributed Systems \& Java/Go: Projects like apache/pulsar and cloudquery/cloudquery offer high-complexity scenarios involving concurrency and system architecture.
    \item Modern Tooling \& Rust/C: Repositories like astral-sh/ruff and tursodatabase/libsql represent the modern shift towards high-performance systems programming.
\end{itemize}

\section{Efficacy of Rejection Sampling}
\begin{table}[ht]
\centering
\resizebox{\textwidth}{!}{
\begin{tabular}{cccccccc}
\toprule
\textbf{Model}    & \textbf{Query} & \textbf{SWE-Bench}            & \textbf{Toolathlon} & \textbf{DS-1000}              & \textbf{\makecell{$\tau^2$-bench\\-retail}} & \textbf{SciCode-MP} & \textbf{AVG} \\
\midrule
\textbf{GLM-4.6}  & -              & 0.608                         & 0.157               & 0.522                         & 0.675                      & 0.062               & 0.405      \\
\textbf{+No Rejection Sampling} & 1181           & 0.090                         & 0.000               & 0.322                         & 0.568                      & 0.046               & 0.205  \\
\textbf{+Rejection Sampling} & 1181           & 0.618 & 0.144               & 0.544 & 0.632                      & 0.169               & 0.421 \\
\bottomrule
\end{tabular}

}
\caption{Ablation study on the impact of Rejection Sampling. Results demonstrate that filtering for high-quality trajectories via rejection sampling significantly enhances the fine-tuned model's performance across diverse benchmarks compared to the unfiltered baseline.}
\label{tab:reject_sampling}
\end{table}
To rigorously isolate the impact of data quality on downstream performance, we conducted a controlled ablation study comparing the base GLM-4.6 model against variants fine-tuned with and without our rejection sampling filter. As presented in the accompanying table, training on raw, unfiltered trajectories leads to a catastrophic performance degradation, with the average score plummeting to 0.205—significantly below the base model's 0.405. This sharp decline demonstrates that indiscriminately ingesting self-generated data introduces substantial noise, causing the model to effectively unlearn its pre-trained capabilities and validating the hypothesis that low-quality supervision is detrimental to agentic reasoning.

In stark contrast, implementing the rejection sampling mechanism reverses this trend, not only recovering the model's baseline capabilities but enhancing them to a superior average score of 0.421. The benefit of this high-fidelity supervision is particularly pronounced in complex, domain-specific tasks; for instance, performance on the rigorous SciCode-MP benchmark more than doubles, rising from 0.062 in the base model to 0.169. These findings compellingly prove that the semantic alignment ensured by rejection sampling is the decisive factor in successful self-distillation, allowing the model to improve through valid reasoning paths while filtering out hallucinatory logic.

\section{Query Constructor Prompt}

\begin{tcolorbox}[title={Query Constructor Prompt}, 
                  colback=blue!5!white, 
                  colframe=blue!75!black,
                  breakable]

\textbf{\#\# Role}\\
You are an AI Model Test Case Designer. Your expertise lies in reverse-engineering code changes to create an ``intent-based query" used for benchmarking AI code generation models.\\

\textbf{\#\# Core Task}\\
Your task is to analyze the provided Pull Request (PR) data and generate an intent-based query.\\

This query serves one purpose: It will be the input prompt for a separate AI coding agent. The test is to benchmark if that agent can reproduce the functional changes described in the original PR data.\\

\textbf{\#\# Key Rules \& Constraints}\\

\textbf{\#\#\# Core Thought Process }\\
You must follow this 4-step analysis to generate the query:\\
- Identify Core Intent (The ``Why"): Analyze the PR title and description to find the fundamental problem being solved or the strategic goal.\\
- Pinpoint Conceptual Locus (The ``Where"): Examine the patch to identify the modified system area (files or classes). You must inform the file paths to be executed and conceptually explain the variables involved.\\
- Abstract Key Logic (The ``How"): Analyze the patch and execution plan. Abstract the key logic (e.g., variables, conditional logic) into a ``vague plan," noting only the logical associations between functional areas.\\
- Synthesize Directive (The ``Plan"): Combine the "Why," "Where," and "How" into a single, high-level narrative. Start with the goal (Step 1), guide the AI to the correct conceptual area (Step 2), and explain the required functional change using the abstracted logic (Step 3). This plan must focus purely on intent.\\

\textbf{\#\#\# Key Constraints}\\
The query you generate MUST be a ``conceptual reverse-instruction".\\
- It MUST capture the core "why" (strategic intent or functional goal) behind the PR, not the "what" (the specific code changes).\\
- It MUST contain enough conceptual information to guide the AI to a correct implementation.\\
- It MUST specify file paths for execution and explain the purpose (not the literal name) of variables, but it cannot include detailed execution instructions.\\

\textbf{\#\#\# Strictly Prohibited }\\
- DO NOT generate step-by-step instructions, operational lists, or bullet points.\\
- DO NOT mention literal function names (e.g., calculate\_cost()), class names, or variable names from the code.\\
- DO NOT include specific URLs or code snippets.\\

\textbf{\#\#\# Example}\\
\textbf{\#\#\#\# Input:} \\
- Repository: microsoft/autogen\\
- PR Title: Refined the \texttt{user\_proxy} description.\\
- PR Description: \#\# Why are these changes needed?\\

Refined the user\_proxy description in cases where it is fully automated. Too often, other agents assumed it could answer on behalf of a user in human\_input=NEVER mode, but it cannot.\\

- Patch JSON: [{``path": "autogen/agentchat/user\_proxy\_agent.py", ``patch": "@@ -20,7 +20,7 @@ class UserProxyAgent(ConversableAgent):     DEFAULT\_USER\_PROXY\_AGENT\_DESCRIPTIONS = {         ``ALWAYS": ``An attentive HUMAN user who can answer questions about the task, and can perform tasks such as running Python code or inputting command line commands at a Linux terminal and reporting back the execution results.",         ``TERMINATE": ``A user that can run Python code or input command line commands at a Linux terminal and report back the execution results.",-        ``NEVER": ``A user that can run Python code or input command line commands at a Linux terminal and report back the execution results.",+        ``NEVER": ``A user that CANNOT WRITE CODE, or answer other questions, but CAN EXECUTE PYTHON CODE and CAN INPUT COMMAND LINE sh COMMANDS at a Linux terminal, and REPORT BACK THE EXECUTION RESULTS.",     }      def \_\_init\_\_("}, {``path": ``autogen/agentchat/user\_proxy\_agent.py", ``patch": "@@ -20,7 +20,7 @@ class UserProxyAgent(ConversableAgent):     DEFAULT\_USER\_PROXY\_AGENT\_DESCRIPTIONS = {         ``ALWAYS": ``An attentive HUMAN user who can answer questions about the task, and can perform tasks such as running Python code or inputting command line commands at a Linux terminal and reporting back the execution results.",         ``TERMINATE": ``A user that can run Python code or input command line commands at a Linux terminal and report back the execution results.",-        ``NEVER": ``A user that CANNOT WRITE CODE, or answer other questions, but CAN EXECUTE PYTHON CODE and CAN INPUT COMMAND LINE sh COMMANDS at a Linux terminal, and REPORT BACK THE EXECUTION RESULTS.",+        ``NEVER": ``A computer terminal that performs no other action than running Python scripts (provided to it quoted in ```python code blocks), or sh shell scripts (provided to it quoted in ```sh code blocks).",     }      def \_\_init\_\_("}]\\
- Specific Execute Plan: I need to update the user\_proxy agent's description to clearly state that it cannot provide user-level responses when human\_input is set to NEVER, ensuring other agents don't mistakenly rely on it for input in fully automated workflows. I'll start by modifying the docstring and internal prompt in user\_proxy\_agent.py to reflect this limitation more explicitly, based on recent team discussions. The main challenge will be phrasing the message clearly without introducing ambiguity, so I'll prioritize precision and consistency with our agent communication patterns.\\

\textbf{\#\#\#\# Output: }\\
In a multi-agent system, it is crucial to prevent other agents from misinterpreting a fully automatic UserProxyAgent as a conversational entity, such as a human or an LLM, especially when its human\_input\_mode is set to 'NEVER'. In this mode, the agent's sole function is code execution. Your task is to examine the operational logic within autogen/agentchat/user\_proxy\_agent.py. Focus on the UserProxyAgent's initialization process to identify and adjust the key variable storing its default descriptions. You must modify the default description specifically for the 'NEVER' mode to unambiguously define its role as a ``code terminal," making it clear that it only executes code and cannot process invalid natural language requests, such as seeking opinions or confirmations.\\

\textbf{\#\# Pull Request (PR) data}\\
- Repository: \{REPO\_NAME\}\\
- PR Title: \{PR\_TITLE\}\\
- PR Description: \{PR\_DESCRIPTION\}\\
- Patch JSON: \{PATCH\_JSON\}\\
- Specific Execute Plan: \{REASONING\}\\

\textbf{\#\# Output Format}\\
- Your response MUST ONLY contain the generated `query' string. \\
- Your answer MUST be in ENGLISH.\\
- Do not include any additional explanations, apologies, or preambles (e.g., ``Here is the query you requested...").\\

\end{tcolorbox}

\section{Evaluator Prompt}
\begin{tcolorbox}[title={Evaluator Prompt}, 
                  colback=blue!5!white, 
                  colframe=blue!75!black,
                  breakable]

\textbf{\# Code Model Judge Prompt}\\

You are an expert code reviewer tasked with evaluating whether an AI-generated patch achieves the same functional effect as a reference GitHub PR patch. Your role is to identify errors, bugs, and deviations from requirements while recognizing that different implementations can be functionally equivalent.

\textbf{\#\# Input Format}\\

You will receive the following inputs:\\

**Original Ground-Truth GitHub PR patch:**\\

\{\{ground\_truth\_patch\}\}\\

\textbf{**AI-generated patch:**}\\

\{\{diff\_patch\}\}\\

\textbf{**User query:**}\\

\{\{test\_query\}\}\\

\textbf{\#\# Evaluation Criteria}\\

\textbf{\#\#\# What to Consider as CORRECT}\\
- **Functional equivalence**: Different implementation approaches that achieve the same end result\\
- **Code style variations**: Different variable names, code organization, or formatting\\
- **Enhanced implementations**: AI adds extra features or improvements not explicitly forbidden\\
- **Alternative logic structures**: Different conditional flows that produce identical outcomes\\
- **More comprehensive tests**: AI writes additional test cases beyond minimum requirements\\

\textbf{\#\#\# What to Flag as ERRORS}\\

\textbf{\#\#\#\# 1. **Critical Bugs**}\\
- Runtime errors (infinite recursion, null pointer exceptions, etc.)\\
- Logic errors that would cause incorrect behavior\\
- Security vulnerabilities\\
- Performance issues that significantly impact functionality\\

\textbf{\#\#\#\# 2. **Requirement Violations**}\\
- Missing required functionality explicitly stated in the user query\\
- Implementing opposite behavior from what was requested\\
- Ignoring explicit constraints or specifications\\
- Breaking existing functionality that should be preserved\\

\textbf{\#\#\#\# 3. **API Contract Violations**}\\
- Wrong method signatures when specifically required\\
- Missing required return values\\
- Incorrect exception types when specified\\
- Breaking changes to public interfaces\\

\textbf{\#\# Analysis Framework}\\

\textbf{\#\#\# Step 1: Functional Correctness Analysis}\\
1. **Core functionality**: Does the AI implementation achieve the primary objective?\\
2. **Edge cases**: Are boundary conditions and error scenarios handled correctly?\\
3. **Integration**: Does it work correctly with existing code?\\
4. **Side effects**: Are there unintended consequences?\\

\textbf{\#\#\# Step 2: Requirement Compliance Check}\\
1. **Explicit requirements**: Check each requirement from the user query\\
2. **Implicit requirements**: Consider standard practices and conventions\\
3. **Constraints**: Verify all limitations and restrictions are respected\\
4. **Test coverage**: Ensure all specified test scenarios are implemented\\

\textbf{\#\#\# Step 3: Bug Detection}\\
1. **Syntax errors**: Check for compilation/parsing issues\\
2. **Runtime errors**: Look for potential crashes or exceptions\\
3. **Logic errors**: Verify algorithmic correctness\\
4. **Resource issues**: Check for memory leaks, infinite loops, etc.\\

\textbf{\#\# Final Response Format}\\

Return a **strict JSON object** with the following shape (no additional keys or prose):\\

json\\
\{
  ``score": 0.0,\\
  ``comment": "Short summary that references the most critical findings or explicitly states the patch is acceptable."\\
\}\\

- `score' must be a floating-point number between **0.0** and **1.0**. Use **1.0** only when the patch is acceptable with no critical errors; use values close to **0.0** when you discover blocking issues.\\
- `comment' must be a concise natural-language explanation describing your overall verdict (e.g., highlight the key blocking issue or state that the patch aligns with the ground truth).\\

Do **not** wrap the JSON in Markdown fences, and do **not** include any additional narrative outside the JSON object.

\end{tcolorbox}

\section{Training Parameters}

\begin{tcolorbox}[title={Training Parameters}, 
                  colback=blue!5!white, 
                  colframe=blue!75!black,
                  breakable]

SFT\_ARGS=(\\
   \texttt{--}rollout-function-path slime.rollout.sft\_rollout.generate\_rollout\\
   \texttt{--}prompt-data \${input\_path}\\
   \texttt{--}input-key messages\\
   \texttt{--}tool-key tools\\
   \texttt{--}rollout-shuffle\\
   \texttt{--}num-epoch 2\\
   \texttt{--}rollout-batch-size 64\\
   \texttt{--}global-batch-size 64\\

   \texttt{--}loss-type sft\_loss\\
   \texttt{--}calculate-per-token-loss\\
   \texttt{--}disable-compute-advantages-and-returns\\
   \texttt{--}debug-train-only\\
)\\

PERF\_ARGS=(\\
   \texttt{--}tensor-model-parallel-size 8\\
   \texttt{--}sequence-parallel\\
   \texttt{--}pipeline-model-parallel-size 4\\
   \texttt{--}context-parallel-size 2\\
   \texttt{--}expert-model-parallel-size 16\\
   \texttt{--}expert-tensor-parallel-size 1\\

   \texttt{--}recompute-granularity full\\
   \texttt{--}recompute-method uniform\\
   \texttt{--}recompute-num-layers 1\\
   \texttt{--}use-dynamic-batch-size\\
   \texttt{--}max-tokens-per-gpu 65536\\
)\\

OPTIMIZER\_ARGS=(\\
   \texttt{--}optimizer adam\\
   \texttt{--}lr 2e-6\\
   \texttt{--}lr-decay-style cosine\\
   \texttt{--}min-lr 1e-7\\
   \texttt{--}lr-warmup-fraction 0.1\\
   \texttt{--}weight-decay 0.1\\
   \texttt{--}adam-beta1 0.9\\
   \texttt{--}adam-beta2 0.98\\
   \texttt{--}optimizer-cpu-offload\\
   \texttt{--}overlap-cpu-optimizer-d2h-h2d\\
   \texttt{--}use-precision-aware-optimizer\\
)\\

\end{tcolorbox}

\end{document}